\documentclass{article}

\usepackage{times}

\usepackage{amsfonts}
\usepackage{amsmath}
\usepackage{bm}

\usepackage{graphicx}
\usepackage{subfigure}
\usepackage{multirow}

\usepackage{natbib}

\usepackage{hyperref}

\usepackage[accepted]{brain}

\usepackage{appendix}

\icmltitlerunning{A machine learning model for identifying cyclic alternating patterns in the sleeping brain}

\begin{document}

\twocolumn[
\icmltitle{A machine learning model for identifying cyclic alternating patterns in the sleeping brain}

\title{}

\vskip -0.5in
\author{
	\begin{tabular}[t]{c@{\extracolsep{6em}}c}
		{\bf Aditya Chindhade} & {\bf Abhijeet Alshi} \\
		Carnegie Mellon University & Carnegie Mellon University\\
		\texttt{achindha@andrew.cmu.edu} & \texttt{aalshi@andrew.cmu.edu}
	\\
	\\
	\\
		{\bf Aakash Bhatia} & {\bf Kedar Dabhadkar}  \\
		Carnegie Mellon University & Carnegie Mellon University\\
		\texttt{aakashb1@andrew.cmu.edu} & \texttt{kdabhadk@andrew.cmu.edu}
	\\
	\\
	\\
		{\bf Pranav Sivadas Menon}  \\
		Carnegie Mellon University \\
		\texttt{psmenon@andrew.cmu.edu}
	\end{tabular}
}

\date{}
\maketitle
\thispagestyle{empty}
\vskip 0.1in
]


\newcommand\todo[1]{\textcolor{red}{#1}}

\newcommand{\leg}{\bf}

\printAffiliationsAndNotice{*Supported by Philips. 
Presented at HackAuton, Auton Lab, Carnegie Mellon University}


\begin{abstract} 
Electroencephalography (EEG) is a method to record the electrical signals in the brain. Recognizing the EEG patterns in the sleeping brain gives insights into the sleeping disorders. The dataset uploaded under consideration contains data points associated to numerous physiologies. There are particular patterns associated with the Non-Rapid Eye Movement (NREM) sleep cycle of the brain. This study attempts to generalize the detection of these patterns using a machine learning model. The proposed model uses additional feature engineering to incorporate sequential information for training a classifier to predict the occurrence of Cyclic Alternating Pattern (CAP) sequences in the sleep cycle, which are often associate with sleep disorders. 
\end{abstract} 



\section{Introduction}
\label{Introduction}

The electroencephalogram (EEG) is an important tool to derive important information about an individual’s sleeping activity. Recent studies have been introduced in sleep research based on the nature and quantitation of the sleep micro-structure, taking into account the time structure of phasic EEG events observed during non-REM (NREM) stage and shorter than the standardized scoring epoch (Mariani, 2011)

	The cyclic alternating pattern (CAP) is a periodic EEG activity occurring during NREM sleep. It is characterized by cyclic sequences of cerebral activation (phase A) followed by periods of deactivation (phase B). Phase B separates two successive phase A periods with an interval of less than one minute. A CAP cycle is defined as an ‘A’ phase followed by a ‘B’ phase and at least two CAP cycles are required to form a CAP sequence. CAP is also a marker of sleep instability and can be correlated with several sleep related pathologies. 

One of the studies conducted by (Diego, 2006) aims to analyze and identify the difference in EEG signals between neural patterns of thought and the signals obtained from the imagination of a vowel. Their computation implements the Fourier transform for obtaining frequency data of all the samples, after which ANOVA was used to cross check measurable difference among samples. 

	(Mariani, 2011) analytically evaluated the information content of each description and introduced new descriptors through the application of FFT and normalization of the mean. This helps in improving the automatic recognition of CAP. 

	(Bashivan, 2015) obtained a sequence of topology preserving multi-spectral images as opposed to standard EEG analysis techniques that missed out on the spatial information. After obtaining the EEG movie, a deep recurrent-convolutional neural architecture was trained to learn robust representations from the sequence of images or frame. This approach preserved the spatial, spectral and temporal structure of the EEG data.

	There have been increasing efforts for automatic detection of phase ‘A’ start time and duration using purely data driven approaches. (Novona, 2002)

\subsection{Description of problem statement}

Using various descriptors, the EEG patterns during sleep of 108 participants are obtained. These datasets are publicly available on physionet.org (Terzano, 2002) The participants included 16 healthy subjects and 92 subjects with some past pathological recording, with the following distribution:

\begin{table}[!h]
\begin{center}
    \begin{tabular}{ | p{3cm} | c | c |}
		 
	    \multicolumn{1}{ c }{Pathology} &
	    \multicolumn{1}{ c }{Participants} \\
		\hline\hline
		No pathology & 16\\
		\hline
		Bruxism & 2\\
		\hline
		Insomnia & 9\\
		\hline
		Narcolepsy & 5\\
		\hline
		Nocturnal Frontal Lobe Epilepsy (NFLE) & 40\\
		\hline
		Periodic Leg Movements (PLM) & 10\\
		\hline
		REM behavior disorder (RBD) & 22\\
		\hline
		Sleep disordered breathing (SBD) & 4\\
		
\hline\hline
    \end{tabular}
\caption{{\leg Classification accuracy obtained at various values of Stride and Window size}}
\end{center}
\end{table}
This work is based on the sleep cycle of the first healthy subject who did not have any neurological disorder. The waveforms recorded after every 0.0019531 seconds constitute a total of 17,725,426 data points. Using these data points, feature engineering is performed to form the feature matrix. The feature matrix acts as an input to the logistic regression model, which will account for the time series analysis of the brain signals and classify the start of phase A in a given CAP cycle.

\subsection{Description of data}

There are numerous EEG readings of 108 participants taken recorded after every 0.0019531 seconds. This number varies throughout the data.
	The following is a sample of how the data looks for one participant for the first 40 seconds:

\begin{figure}[!t]
\begin{center}
\centerline{\includegraphics[width=\columnwidth]{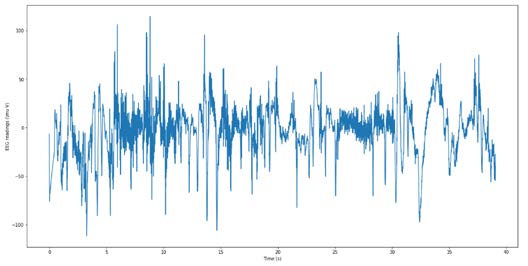}}
	\caption{{\leg Sample data (duration: 40 seconds)}}
\label{fig:goal-driven-1}
\end{center}
\vskip -0.2in
\end{figure}

\section{Methodology}

The data needed for this study is open source  The data is  collected for 108 nights and be to different patients with different psychological conditions. However, for the sake of this study, we only consider the data for a healthy psychology in the first part. After the model is trained on this dataset, it can be extended to a few other datasets to compare performance.
\subsection{Cleaning the data}

\newcommand{\Agent}{\mathcal{A}}
\newcommand{\MDP}{\mathcal{M}}
\newcommand{\Npast}{N_{\mathrm{past}}}

The training data set has 17,725,426 data points, each separated at a fixed time interval of 0.0019531 seconds. The only EEG reading that this study concerns with is F2-F4 signal. Separately, data about the start time and duration of phase A is also available. Both of these data files are combined into a single data file by labelling each data point in the EEG dataset with information about the presence of a CAP cycle. This data file has three columns, column one represents the absolute time (in seconds) at which a reading was recorded, column two is the actual EEG reading (in microvolts), column three, are binary representations (0 or 1), where 0 indicates the  absence of phase A at this given time and 1 indicates presence of phase A at this time. This cleaned data is then used for pre-processing, followed by training a binary classifier.

\subsection{Pre-processing}

The EEG data is highly non-stationary. A non-stationary data is characterized by time-varying mean and variance. It is important to make the data stationary to draw any conclusive evidence relating to the presence of phase A during any given time interval.

Differential moving average is used to make the data stationary. The Dickey-Fuller test (Said, 1984) is used to test stationarity. It was found that a differential moving average with window size of 15 made the data the most stationary, as shown by the corresponding test statistic value. The first window size value was selected randomly. Based on the test statistic value obtained at this point, the next point was selected and so on until the best value was obtained. The original EEG readings were then transformed into differential moving average values using a window size of 15.

\newcommand{\echar}{e_{\mathrm{char}}}

\section{Methodology}
This study intends to classify each data point based on the 0 or 1 value indicated in the third column in the dataset. The objective is to distinguish those data points that fall under the duration of phase A from those that do not. In addition, this dataset is highly sequential. In other words, if phase A occurs at any given data point, the chances of phase A continuing to the next data point are very high. In order to train a classifier for a problem of this type, additional feature 
\subsection{Feature transformation}
The original data points are merely a sequence of stationary EEG values. This sequence can be denoted as follows:

\[X =  [x_1, x_2,  ..x_p,...x_w... x_N ]^T\]

Where, X1, X2 are the first two data points of the sequence, Xw and Xp are some intermediate data points and XN denotes the last data points.

The labels corresponding to these data points can be represented in the following manner-

\[Y =  [y_1, y_2,  ...y_p,...y_w... y_N ]^T\]

yi = 1 indicates the presence of phase A in that duration and yi = 0 indicates the absence.

To ensure sequential properties of the data points are maintained during training, the data points are transformed into data vectors denoted in the following manner-
\[X_1=  [x_1, x_2,  ........................x_w]\]
\[X_2=  [x_p, x_{p+1},  ................x_{w+p}]\]
\[X_{last}=  [x_{n-p}, x_{n-p+1},  ....x_{n-w}]\]

Accordingly, the feauture matrix can be formulated as:
\[
\begin{bmatrix}
    x_{1} & x_{2} & x_{3} & \dots  & x_{w} \\
    x_{p} & x_{p+1} & x_{p+2} & \dots  & x_{w+p} \\
    \vdots & \vdots & \vdots & \ddots & \vdots \\
    x_{n-p} & x_{n-p+1} & x_{n-p+2} & \dots  & x_{n-w}
\end{bmatrix}
\]

where ‘w’ is called the window and ‘p’ is called the stride. Both w and p are hyper-parameters. Their values are determined by running tests on the cross-validation data. It is important to note that if the value of p is greater than w, some data points are lost. So, the upper-limit of p is kept w for cross-validation. 
There are a number of ways in which each of these data vectors can be labelled. A heuristic is to choose the label of the last data point within the vector and assign it to the entire data vector. Another approach could be to method is to implement a maximum vote strategy. In this method, the label is given to any data vector Xi is ‘0’ if maximum number of data points contained in this data vector have label ‘0’, else it is 1. If w is even and there are equal number of ‘0’ and ‘1’ labels to the data points, ‘0’ is chosen as the tie-breaker, since the probability of getting a ‘0’ is much more than the probability of getting a ‘1’.

	There can be combinations of weighted averages of data point labels that could be used to assign labels to a vector. But, since the number of zeros is much greater than the number of ones, it might make sense to assign a vector a label of ‘1’ even if only one data point in the vector has a label of ‘1’. There is a compromise with detecting the accurate duration of phase A, but for the sake of this study, this compromise is ignored, as long as the size of window, w, is fairly small. In the current implementation, we have assigned the value of the last data point to the vector.

\subsection{Logistic Regression}

Logistic regression is implemented on Python using the scikit-learn package (Pedregosa, 2011, Buitinck, 2013).The entire dataset is split into training and test sets in the ratio 70:30. An L2 regularizer was used with a regularization coefficient of 0.01. The optimizer used was ‘Newton-conjugate gradient’ to incorporate parallelization across CPU cores. In order to account for the class imbalance, the weights were balanced using corresponding class frequencies. 
A series of experiments were performed (Fig 2.) as discussed in Section 3.1 to obtain the optimum values of the hyper-parameters. All experiments were performed on Intel Core I-7 machines with 16 GB RAM and 2.90 GHz clock frequency. 

\begin{table}[!h]
\begin{center}
    \begin{tabular}{ | p{3cm} | c | c |}
		 
	    \multicolumn{1}{ c }{Stride (P)} &
	    \multicolumn{1}{ c }{Window (W)} &
	    \multicolumn{1}{ c }{Accuracy \%} \\
	   
	    \hline\hline
	    58 & 512 & 51 \\ 
	   
	    \hline
	    58 & 1024 & 52 \\ 
	    \hline 
		
	    58 & 1536 & 52 \\ 
		\hline
		116 & 512 & 53 \\
		\hline
		116 & 1024 & 53 \\ 
		\hline
		116 & 1536 & 54 \\
		\hline
		479 & 512 & 53\\
		\hline
		479 & 1024 & 54\\
		\hline
		479 & 1536 & 55\\
		\hline
		958 & 1024 & 56\\
		\hline
		958  & 1536 & 57\\
		\hline
		1276 & 1536 & 58\\
		\hline\hline
    \end{tabular}
\caption{{\leg Classification accuracy obtained at various values of Stride and Window size}}
\end{center}
\end{table}
\section{Results and Discussion}
For hyper-parameter tuning, the model is run on test sets with a combination of values of stride p and window w. The lowest time for which any phase A exists in this dataset is 2 seconds. The value of w chosen should be lesser than this value to ensure no occurrence is skipped. The number of time intervals which make up one second intervals is 512. Therefore, the value of w is restricted to multiples of w. whereas, to ensure that the value of p stays lesser than w and at the same no data is lost, there are a finite number of values of p to choose from. The results are shown in figure 2. The values of w and p which give the maximum classification accuracy (58\%) are chosen. In this case, the optimum values of p and w are 1276 and 1536 respectively.
The Area-Under-Curve Receiver Operating Characteristic (AUROC) value for the optimum combination (1276 and 1536) comes out to 0.512.
The results plotted in terms of the Receiver Operating Characteristic (ROC) curve are shown in figure 3.

\begin{figure}[!t]
\begin{center}
\centerline{\includegraphics[width=\columnwidth]{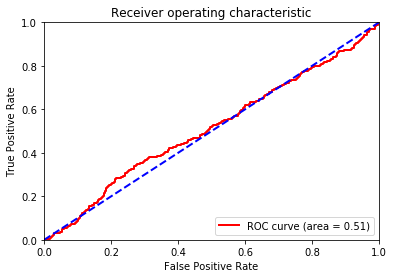}}
	\caption{{\leg Area under the Receiver Operating Characteristic curve for test data}}
\label{fig:goal-driven-1}
\end{center}
\vskip -0.2in
\end{figure}

\section{Conclusion and Future work}
A simple binary logistic regression classifier was trained for classifying EEG data in the sleep cycle into phase A and non-phase A. This study attempted to train raw time-domain data as recorded by the EEG directly without frequency-domain transform. Such a model can be trained and interpreted easily. In addition, this also stresses the importance of using sequential data models like Markov chains and recurrent neural networks on time-domain data. However, considerable accuracy was not obtained using this model. Further work with sequential models might provide better insights.

\section{Acknowledgements}
We thank the Auton Lab and Professor Artur Dubrawski for providing a platform for showcasing this research. We thank Philips for motivating the problem statement and we also thank Pittsburgh Supercomputing Centre for allowing us to access their computational resources.

\section{References}
\begin{enumerate}
\item Mariani, Sara, et al. "Characterization of A phases during the cyclic alternating pattern of sleep." Clinical Neurophysiology 122.10 (2011): 2016-2024.
\item Rojas, Diego Alfonso, Leonardo Andrés Góngora, and Olga Lucia Ramos. "EEG signal analysis related to speech process through BCI device EMOTIV, FFT and statistical methods." (2006).
\item Bashivan, Pouya, et al. "Learning representations from EEG with deep recurrent-convolutional neural networks." arXiv preprint arXiv:1511.06448 (2015).
 \item Navona, Carlo, et al. "An automatic method for the recognition and classification of the A-phases of the cyclic alternating pattern." Clinical neurophysiology 113.11 (2002): 1826-1831.

\item Terzano, Mario Giovanni, et al. "Atlas, rules, and recording techniques for the scoring of cyclic alternating pattern (CAP) in human sleep." Sleep medicine 3.2 (2002): 187-199.

\item Said, Said E., and David A. Dickey. "Testing for unit roots in autoregressive-moving average models of unknown order." Biometrika 71.3 (1984): 599-607.
\item Pedregosa, Fabian, et al. "Scikit-learn: Machine learning in Python." Journal of machine learning research 12.Oct (2011): 2825-2830.
\item Buitinck, Lars, et al. "API design for machine learning software: experiences from the scikit-learn project." arXiv preprint arXiv:1309.0238 (2013).

\end{enumerate}

\end{document}